\documentclass[english]{article}
\usepackage[T1]{fontenc}
\usepackage[latin9]{inputenc}
\usepackage{geometry}
\usepackage{array}
\usepackage{longtable}
\usepackage{calc}
\usepackage{url}
\usepackage{multirow}
\usepackage{amssymb}
\usepackage{graphicx}
\usepackage{setspace}
\usepackage[authoryear]{natbib}
\makeatletter

\providecommand{\tabularnewline}{\\}

\makeatother

\usepackage{babel}
\begin{document}
\title{Program synthesis performance constrained by non-linear spatial relations
in Synthetic Visual Reasoning Test}
\author{Lu Yihe\textsuperscript{1}, Scott C. Lowe\textsuperscript{2}, Penelope
A. Lewis\textsuperscript{3}, Mark C. W. van Rossum\textsuperscript{1}}
\maketitle
\begin{center}
{\small{}}\textsuperscript{{\small{}1}}{\small{}School of Psychology,
University of Nottingham, Nottingham, United Kingdom}{\small\par}
\par\end{center}

\begin{center}
{\small{}}\textsuperscript{{\small{}2}}{\small{}Faculty of Computer
Science, Dalhousie University, Halifax and Vector Institute, Toronto,
Canada}{\small\par}
\par\end{center}

\begin{center}
{\small{}}\textsuperscript{{\small{}3}}{\small{}School of Psychology,
Cardiff University, Cardiff, United Kingdom}{\small\par}
\par\end{center}
\begin{abstract}
Despite remarkable advances in automated visual recognition by machines,
some visual tasks remain challenging for machines. \citet{fleuret2011comparing}
introduced the Synthetic Visual Reasoning Test (SVRT) to highlight
this point, which required classification of images consisting of
randomly generated shapes based on hidden abstract rules using only
a few examples. \citet{ellis2015unsupervised} demonstrated that a
program synthesis approach could solve some of the SVRT problems with
unsupervised, few-shot learning, whereas they remained challenging
for several convolutional neural networks trained with thousands of
examples. Here we re-considered the human and machine experiments,
because they followed different protocols and yielded different statistics.
We thus proposed a quantitative reintepretation of the data between
the protocols, so that we could make fair comparison between human
and machine performance. We improved the program synthesis classifier
by correcting the image parsings, and compared the results to the
performance of other machine agents and human subjects. We grouped
the SVRT problems into different types by the two aspects of the core
characteristics for classification: shape specification and location
relation. We found that the program synthesis classifier could not
solve problems involving shape distances, because it relied on symbolic
computation which scales poorly with input dimension and adding distances
into such computation would increase the dimension combinatorially
with the number of shapes in an image. Therefore, although the program
synthesis classifier is capable of abstract reasoning, its performance
is highly constrained by the accessible information in image parsings.
\end{abstract}

\section{Introduction}

Progress in visual recognition by machine has been impressive due
to the remarkable development of machine learning (ML) in the recent
decade. However, it has been argued that machines can only be successful
in certain tasks, while they fail to achieve human-like performance
in others. To highlight the difference between machine and human intelligence,
\citet{fleuret2011comparing} introduced the Synthetic Visual Reasoning
Test (SVRT), consisting of 23 image classification problems. All the
images are composed of randomly generated shapes that are simple closed
contours without intersections. In each classification problem there
are two categories of such images. When presented with a new image
in a particular problem, a human subject or a machine agent has to
classify it according to only the previous images and their categories
that have been seen. A hidden abstract rule determines whether the
images belong to the same category (see Table \ref{tab-Example-SVRT}
for example images and classification rules).

\begin{table}
\centering
\includegraphics[width=1\textwidth]{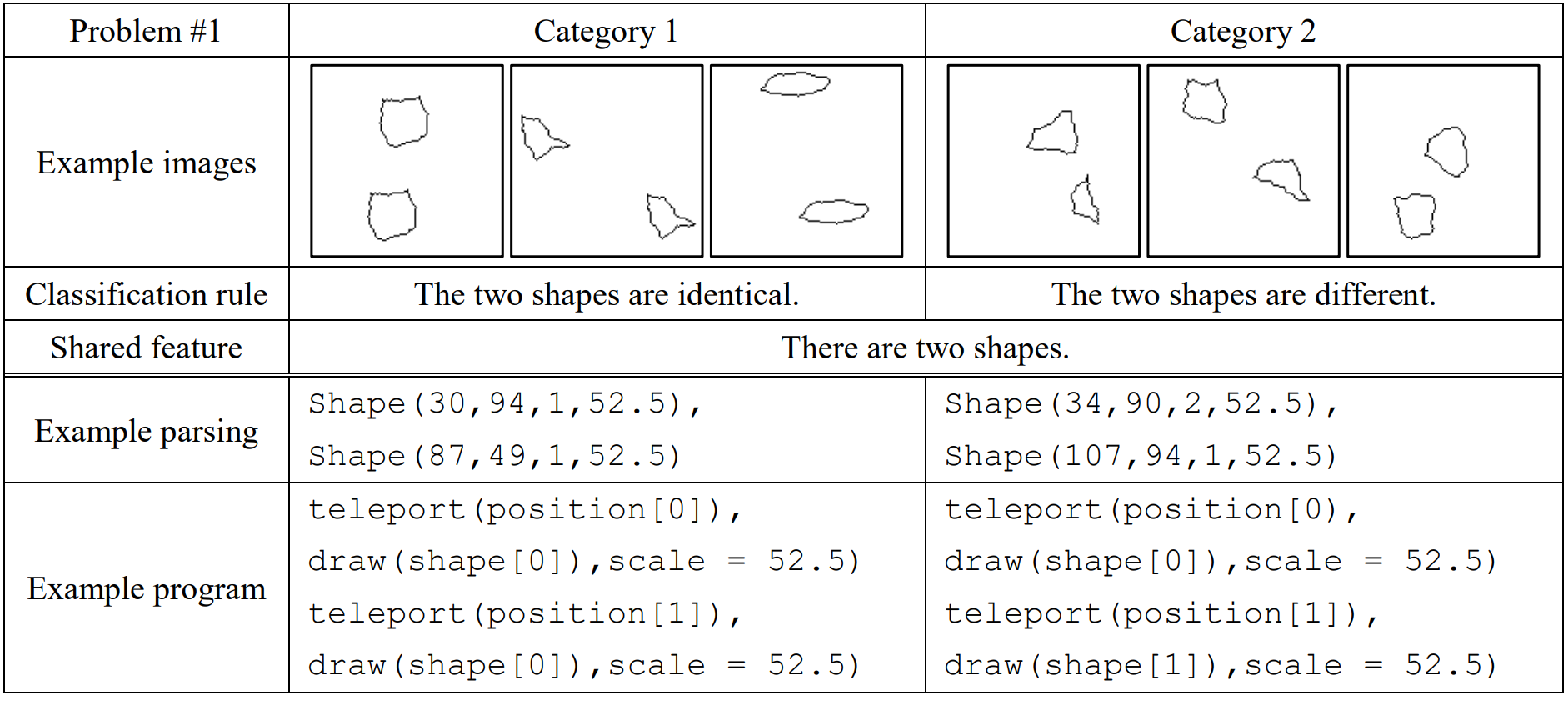}
\caption{\label{tab-Example-SVRT}Problem \#1 in the original SVRT. Each image
parsing contains several \texttt{Shape(x-coordinate, y-coordinate,
shape identity, scale)}, which describes all the shapes respectively
in an image. If two shapes are in contact or one shape is inside another,
a parsing might further contain \texttt{borders(shape index, shape
index)} or \texttt{contains(shape index, shape index)}, where \texttt{shape
index} is implicitly assigned to each shape in the order as they are
represented in the parsing. The example images and parsings were obtained
by our SVRT generator, forked from the original one \citep{fleuret2011comparing}.
The example programs were synthesized by our PS classifier, forked
from Sasquatch \citep{ellis2015unsupervised}. There are no differences
in image generation per se between the original generator and ours,
while there are modifications in parsing extraction and program synthesis
between the previous approach and ours. Example images and classification
rules of more SVRT problems can be found in Table \ref{tab:Original-23}.}
\end{table}

Since all images are composed of randomly generated contours, they
contain little real-world meanings. Therefore, the hidden classification
rules are designed to be associated with abstract reasoning in visual
recognition. In addition, the random generation of images prevents
the usage of low-level cues or brute-force memory, and allows us to
generate any number of images.

Human subjects outperformed the machine agents: the human subjects
could detect or deduce the hidden rules and thus classify images successfully
after seeing and classifying a handful of images (about 6 for most
of the 23 problems), while Support Vector Machine (SVM) and Adaptive
Boosting (AdaBoost) only achieved reasonably high accuracy after being
trained with 10,000 examples \citep{fleuret2011comparing}. Subsequent
studies \citep{ellis2015unsupervised,stabinger201625,ricci2018same}
took the SVRT as a challenge for automated visual recognition, and
employed more advanced ML techniques, in particular Convolutional
Neural Network (CNN) and its variants. Specificially, ConvNet with
2,000 examples \citep{ellis2015unsupervised}, LeNet and GoogLeNet
with 20,000 examples \citep{stabinger201625}, and vanilla CNNs consisting
of different numbers of layers and different sizes of receptive fields
with 2 million examples \citep{ricci2018same}, all failed to achieve
human performance in several problems.

\citet{ricci2018same} sorted the original SVRT problems according
to their CNN performance, and noticed their CNNs could solve problems
involving detection of spatial relations (SR) between shapes, but
not whether two shapes were same or different (SD) in an image. They
further pointed out that the hidden rules of the original 23 problems
were chosen in an arbitrary manner, which made it difficult to analyse
what characeteristics of the problems constrain their CNN performance.
They thus developed the Parametric SVRT (PSVRT), a variant version
of the SVRT in which each image can be characterised by some variability
parameters (e.g., number of shapes), so that the classification difficulty
of a problem (in terms of machine performance) could be correlated
to these image variability parameters. They found that only the SD
problems led to relatively low performance of their CNNs; especially
when image sizes were large, the CNNs needed more than 10 million
examples to reach a reasonably high classification accuracy in the
SD problems.

However, the general conclusions was criticised by \citet{borowski2019notorious}
that all CNNs are unable to solve some difficult SVRT problems and
thus a feedforward convolutional architecture is incapable of corresponding
abstract reasoning, because a poor machine performance might not be
caused by the architecture, but by sub-optimal parameter choices or
training strategies. In fact, they showed that their ResNet50 could
reach over 90\% test accuracy on all the original SVRT problems \citep{borowski2019notorious}.

To achieve an unsupervised, few-shot learning machine performance,
\citet{ellis2015unsupervised} explored the possibility to solve the
SVRT via the program synthesis (PS) approach. The PS approach is generative;
it aims to construct a program that generates the images of a given
class. As the PS approach scales extremely poorly with input dimension,
images were first parsed to yield compact descriptions, i.e. image
parsings. These image parsings were then send to the synthesizer which
contructed an optimal program for each image category. To classify
test images for a problem, a pair of such programs (for the two categories
respectively) were deployed to check which one was more compatible
with the parsing of each test image.

Here we re-considered the experiments and results, as the human and
machine experiments were conducted with different protocols, and different
statistics were used for comparison between human and machine performance.
We were concerned more on the PS approach than the neural approach
due to our interest in its somewhat human-like performance (capable
of unsupervised, few-shot learning). Moreover, we aimed to investigate
the PS approach in details, particularly whether the program contruction
by the synthesizer or simply the problem reduction by the parser played
a crucial role in solving the SVRT. Although \citet{ellis2015unsupervised}
showed that the classification based on the parsings did not lead
to human-like performance when they were used as inputs for AdaBoost,
it did not necessarily imply, as \citet{borowski2019notorious} pointed
out, that the method is unable to solve the SVRT by design, and AdaBoost
was employed for baseline machine performance after all \citep{fleuret2011comparing,ellis2015unsupervised}.
Similar to the previous studies deepening our understanding in different
CNNs by focusing on the difficult problems \citep{stabinger201625,ricci2018same,borowski2019notorious},
we aimed to learn limitations of the PS approach and thanks to its
interpretability we could analyse relevant problems, find out possible
reasons and suggest possible solutions.

In Section \ref{sec:methods}, we will discuss the protocols of the
machine and human experiments and how to interpret and compare their
results despite protocol differences. In Section \ref{sec:Results},
we will investigate the results of our machine experiments, highlighting
the improvements in the PS performance by correcting image parsings,
comparing them to the human and machine results reported by the other
studies with respect to different problem types. In Section \ref{sec:Discussion},
we will summarise the advantages and disadvantages of the PS approach
and discuss the reasons and further improvements might be made in
automated visual recognition.

\section{Materials and methods\label{sec:methods}}

\subsection{Using program synthesis on SVRT\label{subsec:Generating-SVRT}}

\citet{ellis2015unsupervised} demonstrated that their machine agent,
namely Sasquatch, was able to achieve human-like performance on the
SVRT in the sense that it could perform unsupervised, few-shot learning
on many SVRT problems. Sasquatch classifies SVRT images in three steps.
Firstly, its parser pre-processes the raw images and encodes them
into low-dimensional parsings. A parsing of an image contains only
the information of individual shapes and their spatial relations;
specifically, the two-dimensional coordinates of the centre of mass,
the identity and the scale of every shape and whether a shape is bordering
to or contained by another. If two shapes in an image are identical
under any geometric similarity transforms (i.e., translation, rotation,
rescaling and reflection), they share the same identity. The normalised
sizes of shapes that share the same identity in an image are considered
to be their scales, with the scale of the largest shape set to $1$;
if two shapes do not share the same identity, it is meaningless to
compare their scales. Secondly, the Sasquatch synthesizer constructs
an optimal program for each image category by training on 3 examples.
As each problem contains two categories (which we call positive and
negative categories), it constructs two programs respectively for
them with 6 training examples in total. Finally, when classifying
a test image, the evaluator computes how compatible an image is with
either program by some likelihood function and classifies the image
accordingly (see Table \ref{tab-Example-SVRT} for example parsings
and programs). In the second and third steps, it employs a theorem
prover for general formal verification, Z3 (which is made public by
\citet{Z3} at \url{https://github.com/Z3Prover/}), to check the
satisfiability of constructed programs and test images. In other words,
Sasquatch attempted to solve the SVRT as a Satisfiability Modulo Theory
(SMT) problem. Since an SMT problem is proved to be NP-complete \citep{cook1971complexity},
verifying a program requires time that is superpolynomial in the input
dimension. Therefore, a fixed time limit was set for Z3 in each verification
iteration. In total, the construction of a single optimal program
took from around 0.05 seconds to more than 2 minutes depending on
the problems (running on an individual iMac with a 2.7 GHz Intel Core
i5 processor).

However, Sasquatch did not fully solve the SVRT. We noticed that some
of its poor performance was caused by incorrect parsings. In particular,
Sasquatch did not assign the same shape identity to identical shapes
if rotated, rescaled or reflected (unless the rotation or rescaling
is trivially small); it could tell which shape was bigger if two shapes
were different; and occasionally when several shapes were close to
one another it parsed the space surrounded by them as an extra shape.
In order to test the impact of such image parsing errors on classification
performance, we obtained our corrected parsings by extracting shape
information directly from the SVRT generator, and then conducted experiments
on both sets of parsings. In particular, our parsings contain the
correct identities of every shape and the accurate coordinates of
the generation centre (which is the point for generating random contour
around it, not the centre of mass). Moreover, we re-define their scales
to be the real sizes rather than the normalised ones, and a negative
scale was used if a shape is a reflected copy of another. Our code
can be found at \url{https://github.com/anish-lu-yihe/pySVRT}. In
addition, rotated angles of similar shapes can be encoded in our new
parsings as an extra feature. However, we did not use this feature
for two reasons: firstly, the synthesizer scales poorly with its input
dimension; secondly, rotation is not a necessary feature for classifying
any image in the 23 SVRT problems.

Since we do not changed the synthesizer and the evaluator, except
for the parameters relevant to the corrections in parsings, our machine
agent remains a PS classifier, whose code can be found at \url{https://github.com/anish-lu-yihe/SVRT-by-Sasquatch}.
We also used AdaBoost for a baseline machine performance \citep{fleuret2011comparing,ellis2015unsupervised}.
The input vectors were taken directly from the image parsing by ripping
off the non-numeric tags, and the outputs are binary. In particular,
we employed the AdaBoost classifier of Scikit-learn \citep{scikit-learn}.
We found that there were little differences in the results across
different numbers of stumps and thus report only the performance of
AdaBoost with 10,000 stumps in Table \ref{tab:Adaboost-performance}.
The code and results can be found at \url{https://github.com/anish-lu-yihe/SVRT-by-AdaBoost}.

\subsection{Comparing machine to human performance\label{subsec:Comparing-SVRT}}

Although \citet{fleuret2011comparing} pointed out humans outperformed
machines in the SVRT, it is worth noting that the experimental protocols
of human and machine experiments are different. In the human experiment,
20 subjects were presented with images in a sequence. After making
the classification decision, the subject was informed of the true
category of the test image and a test image was shown while past test
images remained on the display. The subject was considered successful
if 7 correct decisions were made in a row within 35 images. For a
given SVRT task, the number of the succesful subjects and the average
number of images before making 7 correct decisions in a row characterised
the average human performance.

In the machine experiment, an agent was firstly trained with up to
10,000 example images, and then tested on a separate set of images
without further learning. Subsequent analysis of the SVRT followed
the same experimental protocol \citep{ellis2015unsupervised,stabinger201625,ricci2018same,borowski2019notorious}.
We also followed the same protocol when performing our machine experiments.

While for most of the ML techniques thousands or even millions of
examples were used for training, few-shot learning can be realised
by PS. In particular, Sasquatch was trained on 3 pairs of positive
and negative examples for each problem \citep{ellis2015unsupervised},
and for comparison we used the same size for the training sets, except
when we looked into the learning curves by PS. The test set contained
94 unseen images. For each problem, we repeated the same training
and testing procedure with our PS classifier for 40 times, and averaged
the test classification accuracies. Due to the repetition, the classifier
could spend more than 1 hour on some problem (without parallelisation).

The number of successful human subjects and the test accurracies of
machine agents are related but different statistics. For more quantitative
analysis, we denote the machine classification accuracy in a single
test trial by $\alpha$ on one hand. Since a machine can only learn
during the training but have no funtion to update itself further when
being tested, we assume each test trial is independent to one another,
and approximate $\alpha$ by the overall test accuracy. On the other
hand, we denote the proportion of successful human subjects by \textbf{$\beta$},
which is found by dividing the number of successful subjects through
20 (the total number of subjects), using the numbers reported in \citet{fleuret2011comparing}.

\begin{figure}
\centering

\includegraphics[height=10cm]{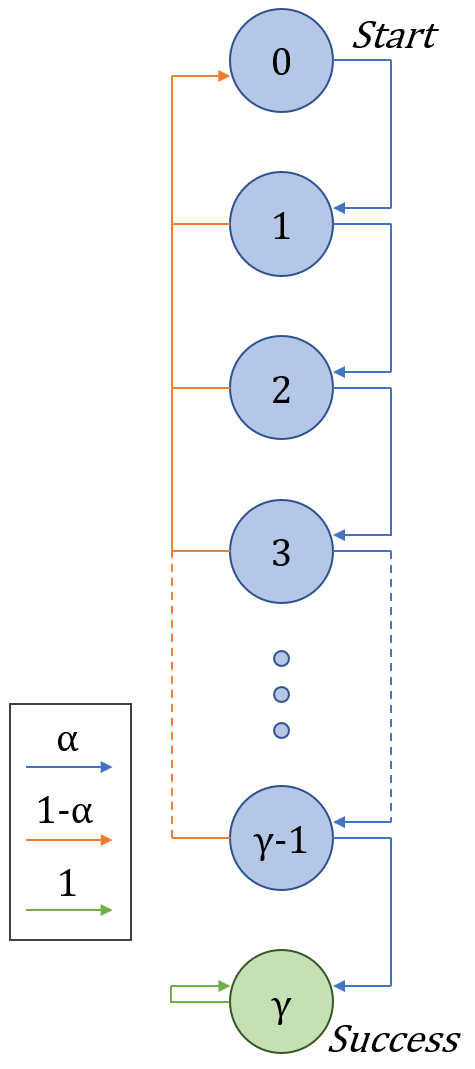}\includegraphics[height=10cm]{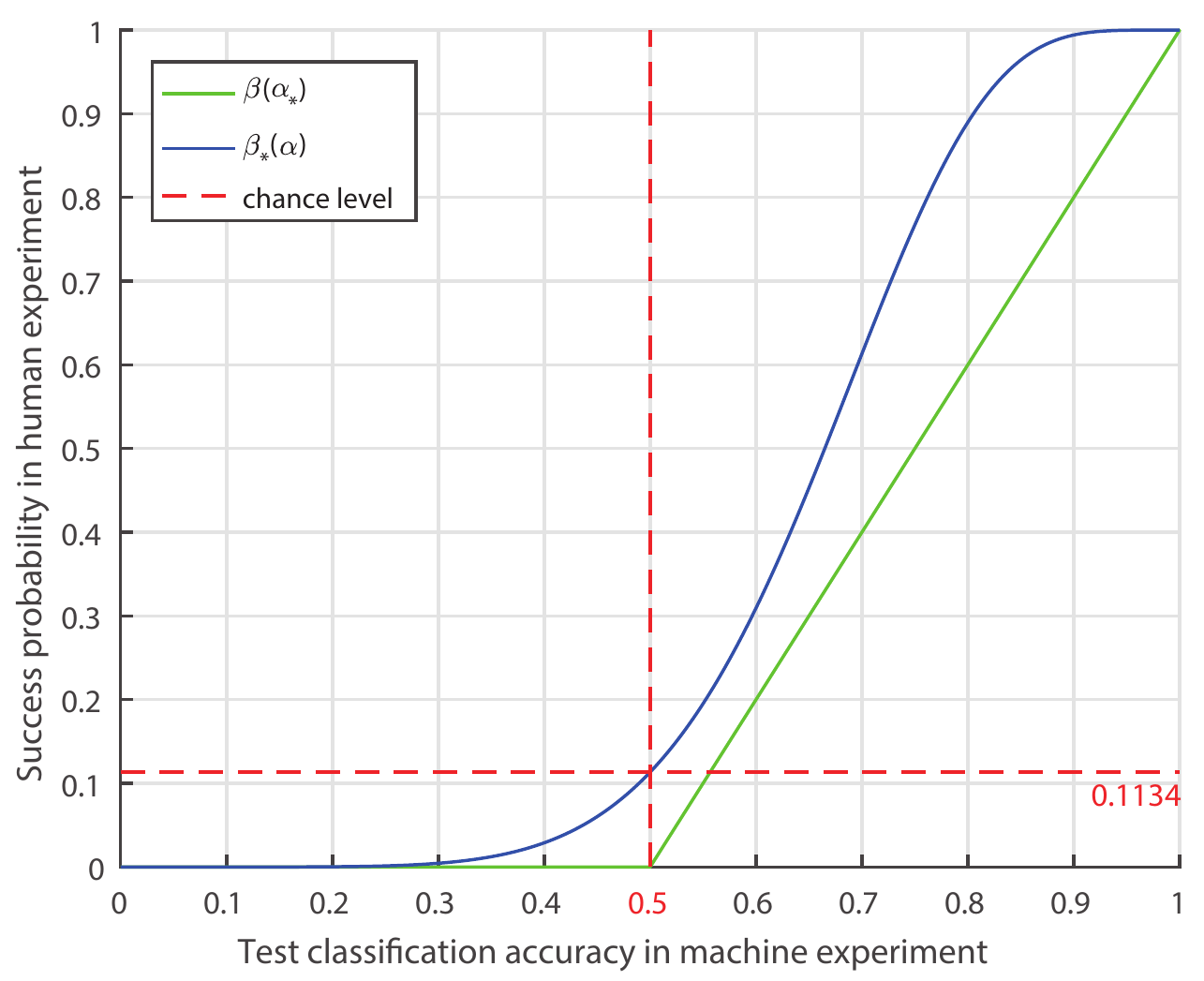}

\caption{\label{fig:markovchain}Reinterpretation of machine performance with
the success probability $\beta_{*}(\alpha)$ for comparison with the
human data. A machine agent of a test classification accuracy $\alpha$
attending a human experiment can be modelled as a discrete Markov
chain (as the diagram on the left), where $Q_{k}(N)$ is the probability
of the $k$ correct classifications in a row at the $N$-th step.
As a success in the human experiment of \citet{fleuret2011comparing}
required 7 correct classifications in a row by seeing up to 35 examples,
$\beta_{*}(\alpha)=Q_{7}(35)$ is the probability of the machine agent
to succeed in the human experiment (the blue curve in the graph on
the right). The detailed derivation can be found in Section \ref{appendix}.
The chance level of $\alpha$ is $50\%$, and that of $\beta_{*}(\alpha)$
is approximately $11.34\%$. \citet{stabinger201625} reinterpreted
the human performance by $\alpha_{*}(\beta)$ with $\beta=[2\alpha_{*}-1]_{+}$
(the green curve in the graph on the right), which effectively overestimated
the human performance and underestimated the machine performance.}
\end{figure}

We consider $\beta$ as the average human performance, overlooking
individual differences. Although both $\alpha$ and $\beta$ indicate
average human and machine performance in percentage, their values
cannot be directly compared due to the different human and machine
experimental protocols. \citet{stabinger201625} reinterpreted the
human performance by $\alpha_{*}(\beta)=(1+\beta)/2$, assuming any
successful subject can perfectly classify all test images while any
unsuccessful subjects make decisions at the chance level (i.e., $50\%$).
In contrast, we reintepret the machine performance by $\beta_{*}(\alpha)$,
the probability for a trained machine to be successful in a human
experiment (see Figure \ref{fig:markovchain}). It is clear to see
that the $\alpha_{*}$ reinterpretation overestimates the human performance
and underestimates the machine performance comparing to our $\beta_{*}$
reinterpretation.

\subsection{Grouping SVRT problems by core characteristics \label{subsec:Grouping-SVRT}}

\citet{fleuret2011comparing} plotted for the 23 problems the mean
number of the trials before a human subject successfully detected
the classification rule and the number of unsuccessful subjects, showing
that there were 4 problems relatively difficult to human. In these
problems, no less than 7 out of the 20 subjects were unsuccessful
($\beta<70\%$), which also took on average more than 12 images for
the successful subjects before detecting the classification rules
while the average is less than 7 for all other problems. In contrast,
AdaBoost yielded a test error over 10\% ($\alpha<90\%$) in 13 out
of the 23 problems, when trained on 10,000 examples. \citet{ellis2015unsupervised}
considered a problem as \textit{solved} by machines if $\alpha\geq90\%$,
calculated the correlations between individual machines and human
performance (i.e., the Pearson's $r$ between $\alpha$ and $\beta$
for each machine agent), and remarked that their PS classifier Sasquatch
was the best machine agent, comparing to AdaBoost and ConvNet, in
terms of the number of solved problems and the correlation to the
human performance.

As pointed out by \citet{ricci2018same}, however, comparing the number
of solved problems is not a good idea for the general comparison between
human and machine performance on SVRT. If a machine agent is able
(or unable) to solve some problem, it is highly likely it would succeed
(or fail) in similar problems. The 23 problems could contain many
easy (or difficult), similar problems for this agent, because the
original 23 problems were arbitrarily designed. Moreover, these relatively
high (or low) machine accuracy might have uncorrected impact on the
Pearson's $r$, which is known to be not robust with respect to outliers,
especially when sample size is small and non-normal; in the statistical
analysis on the correlation between human and machine accuracies by
\citet{ellis2015unsupervised}, there are effectively only 23 data
points. In order to freely control the generation of SVRT-like problems,
\citet{ricci2018same} proposed the PSVRT, so that they could systematically
investigate the correlations between the parameters that modulate
the image variability and the performance of their CNNs. Specifically,
they firstly sorted the original 23 SVRT problems according to their
CNN performances, noticing that their CNNs were good at the SR problems
but poor at the SD problems. Then they developed the PSVRT which could
generate random images based on to two core parameters, shape sizes
(relative to image size) and number of shapes, modulating SR and SD
respectively \citep{ricci2018same}.

Based on the information encoded by image parsings and inspired by
the idea of SD and SR, we propose to group SVRT problems into different
types with respect to their core characteristics for classification,
considering only two aspects of classification features, namely shape
specification (SS) and location relation (LR). In particular, we assume
four levels of complexity in both SS and LR, which are summarised
in Table \ref{tab:Definitions-of-LOC}. With respect to SS, a classification
rule can only rely on the core characteristics whether two shapes
are similar in size ($SS=1$), or whether they are identical ($SS=2$)
or identical under similarity transformations ($SS=3$). Any other
geometric transformation would effectively change shape identities
and reduce identical shapes to completely different shapes, because
all shapes are randomly generated. In such cases, the classification
rule is independent of shape specification ($SS=0$). Thus, the four
levels of SS complexity characterise in principle all possible individual
random shapes.

With respect to LR, a classification rule not based on a trivial,
isolating relation ($LR=0$) either takes shape coordinates into account
($LR=2$ or $3$) or not ($LR=1$, a non-trivial topological relation).
Generally on a two-dimensional plane, there exists only four topological
relations between two shapes of finite area: isolating (the trivial
one), bordering, containing and intersecting \citep{egenhofer1991point}.
As the SVRT generator was designed to produce non-intersecting, random
contours to avoid ambiguity in detecting shape identities, the lowest
two levels of LR ($LR=0$ and $1$) exhaust all possible pairwise
topological relations, which implicitly induce all possible topological
relations among shapes in an image. In the cases when specific shape
coordinates contribute to the core characteristics for classification,
we assume that a non-linear relation in the shape coordinates ($LR=3$)
is harder than a linear one ($LR=2$). This assumption is made based
on the fact that PS processes non-linear numerical expressions much
slower than linear ones, because the underlying algorithm for non-linear
arithmetics is different from and more difficult than linear arithmetics
\citep{Z3}. In addition, a pattern determined by a linear relation
is also seemingly easier for human to detect, as shapes are likely
to form a regular pattern, e.g., aligning in a line or at the corners
of a square. Thus, the four levels of LR cover in principle all possible
location relations among shapes.

In summary, we can group any SVRT problems into $15=4\times4-1$ types
in total according to SS and LR (no problem can have $SS=LR=0$).
In order to solve problems of a particular type, the same type of
features are necessary for any human or machine; we call such information
the \textit{core characteristics} for classification. Since the core
characteristics ignores any features shared by both categories, the
pair of $(SS,LR)$ qualitatively characterises the lowest amount of
information necessary for classification.

\begin{table}
\centering

\begin{tabular}{|c|l|}
\hline
SS & Shape specification \tabularnewline
\hline
0 & All shapes are completely different.\tabularnewline
\hline
1 & Some shapes are not identical but similar in size to one another.\tabularnewline
\hline
2 & Some shapes are identical to one another.\tabularnewline
\hline
3 & Some shapes are identical to one another under similarity transformations
(i.e., scaling, reflection, rotation).\tabularnewline
\hline
\hline
LR & Location relation \tabularnewline
\hline
0 & All shapes have a default, isolating relation.\tabularnewline
\hline
1 & Some shapes have a topological relation (i.e., a bordering or containing
relation).\tabularnewline
\hline
2 & Some shapes have a linear relation in their coordinates (e.g., shapes
aligned in a line/parallelogram).\tabularnewline
\hline
3 & Some shapes have a non-linear relation in their coordinates (e.g.,
equidistances between shapes).\tabularnewline
\hline
\end{tabular}

\caption{\label{tab:Definitions-of-LOC}Definitions of the four levels of complexity
with respect to SS and LR. All shapes are randomly generated contours
with no intersections (there is at least 1 pixel between two contours).
Thus, the default, null relation between two shapes corresponds to
$SS=LR=0$, and a problem cannot belong to this null type because
otherwise no classification rule exists.}
\end{table}
The 23 original SVRT problems can be grouped into 8 types as shown
in Table \ref{tab:Different-types-23}. It is worth noting that the
core characteristics for classification is only necessary but not
sufficient for classification; in other words, although different
problems of the same type require common core characteristics, their
classification rules are not necessarily identical. For example, the
classification rule of problem \#9 depends on whether the location
of a larger shape lies between two smaller ones, while problem \#13
depends on whether the relative locations of a larger shape with respect
to a smaller one in two pairs are identical (i.e., whether there are
two identical meta-shapes, each consisting of a larger and a smaller
one). In addition, SS and LR can be correlated to each other in an
image to some extent, whereas they seem conceptually independent to
each other. For example, detecting whether two shapes have a bordering
relation ($LR=1$) solves problem \#2, while calculating how close
their coordinates are can also be useful ($SS=2$), because the smaller
shape is always inside the larger one and with a high probability
it lies about the centre of the larger shape if not having a bordering
relation. We assume problem \#2 belonging to type $(SS,LR)=(0,1)$
rather than $(SS,LR)=(2,0)$, and the same idea applies and determines
problem types for similar situations. Moreover, there are some disagreements
between our SS-LR grouping and the SD-SR sorting by \citet{ricci2018same},
whereas the idea of SD is similar to SS as they both describe indivdual
shapes. They sorted problem \#6, \#9, \#12, \#13 and \#17 into the
purely LR type, while we consider SS important for solving these 5
problems. For example, solving problem \#9 clearly requires the core
characteristics of shape sizes, which we believe is a characteristic
of individual shapes rather than spatial relations between them.

\begin{table}
\centering

\begin{tabular}{|c|c|c|c|c|c|}
\hline
\multirow{4}{*}{LR} & 3 &  & \#12 & \#6, \#17 & \tabularnewline
\cline{2-6}
 & 2 & \#10, \#14, \#18 & \#9, \#13 &  & \tabularnewline
\cline{2-6}
 & 1 & \#2, \#3, \#4, \#11, \#23 &  & \#8 & \tabularnewline
\cline{2-6}
 & 0 &  &  & \#1, \#5, \#7, \#15, \#19, \#20, \#21, \#22 & \#16\tabularnewline
\hline
\multicolumn{1}{|c}{} &  & 0 & 1 & 2 & 3\tabularnewline
\cline{3-6}
\multicolumn{1}{|c}{} &  & \multicolumn{4}{c|}{SS}\tabularnewline
\hline
\end{tabular}

\caption{\label{tab:Different-types-23}Different types of the original 23
SVRT problems. The problems are grouped by the core characteristics
for classification defined in Table \ref{tab:Definitions-of-LOC}.}
\end{table}

\section{Results\label{sec:Results}}

\subsection{Baseline Adaboost performance with correct parsings}

As pointed out in Section \ref{subsec:Generating-SVRT}, the image
parsings obtained by the Sasquatch parser contained systematic errors
in shape identity. They encoded no information of shape reflection,
which is necessary for problem \#16 and \#20. In problem \#21, rotated
and rescaled shapes were assigned different shape identities unless
the transformations were trivially small. Occassionally in problem
\#2, \#3 and \#11 when the shapes in contact were too close to one
another, the parsings contained tiny shapes that should not exist;
such glitchy shapes were in fact the random spaces well bounded by
other shapes.

\begin{table}
\centering

\begin{tabular}{|c|c|c|c|c|}
\hline
Parsing type & Sasquatch & \multicolumn{2}{c|}{Corrected} & \multirow{3}{*}{Error type in Sasquatch parsings}\tabularnewline
\cline{1-4}
Training size & \multicolumn{2}{c|}{20} & 1000 & \tabularnewline
\cline{1-4}
Test size & \multicolumn{2}{c|}{80} & 1000 & \tabularnewline
\hline
\hline
\#2 & 51.25\% & 57.50\% & 59.90\% & \multirow{3}{*}{Extra glitchy shapes}\tabularnewline
\cline{1-4}
\#3 & 68.75\% & 50.00\% & 51.80\% & \tabularnewline
\cline{1-4}
\#11 & 100.00\% & 100.00\% & 100.00\% & \tabularnewline
\hline
\hline
\#16 & 95.00\% & 100.00\% & 100.00\% & \multirow{3}{*}{Incorrect shape identity}\tabularnewline
\cline{1-4}
\textbf{\#20} & \textbf{50.00\%} & \textbf{100.00\%} & 100.00\% & \tabularnewline
\cline{1-4}
\#21 & 52.50\% & 38.75\% & 47.40\% & \tabularnewline
\hline
\end{tabular}

\caption{\label{tab:Adaboost-performance}AdaBoost performance with Sasquatch
and our corrected parsings. For the problems not listed here, the
Sasquatch parsings are correct, and thus no significant performance
differences can be seen comparing to our parsings. However, even for
the listed problems that the Sasquatch parsings contain errors, only
in problem \#20 the performance difference can be seen.}
\end{table}
Correcting these errors in the parsings, we expected the machine performance
to increase for these problems. However, the performance of AdaBoost
was boosted only in problem \#20 (see Table \ref{tab:Adaboost-performance}).
It is worth noting that, while the Sasquatch parsings failed to serve
AdaBoost well in problems \#20, they led to decent performance in
problem \#16. Although they were blind to shape reflection in both
problem \#16 and \#20, it treated reflected shapes as different shapes
and thus the classifcation rule of problem \#16 changed from `whether
shapes are reflected' to `whether shapes are identical', which coincidentally
resulted in an indifferent performance. Meanwhile, this reflection
blindness resulted in a complete failure in problem \#20, because
the classification rule depends on whether shapes are identical or
not and this information was unretrievable in the Sasquatch parsings.
Since extra glitchy shapes appeared infrequently in problem \#2, \#3
and \#11, the performance were not changed significantly.

Since the number of stumps is essentially the only parameter that
determines the final performance of AdaBoost (whereas there is a tradeoff
in the learning efficiency for the learning rate or the base boosting
algorithm) \citep{scikit-learn}, we conducted the experiments for
10, 100, 1,000 or 10,000 stumps and found the results extremely similar
except for 10 stumps (and thus reported only the results for 10,000
stumps in Table \ref{tab:Adaboost-performance}). Both Sasquatch and
our parsings are in general poor higher-level representations for
AdaBoost; in fact, it yielded worse performance on such parsings than
on features extracted directly from raw images \citep{fleuret2011comparing,ellis2015unsupervised}.

\subsection{Improved PS performance with corrected parsings\label{subsec:Improved-PS-performance}}

\begin{figure}
\includegraphics[width=1\textwidth]{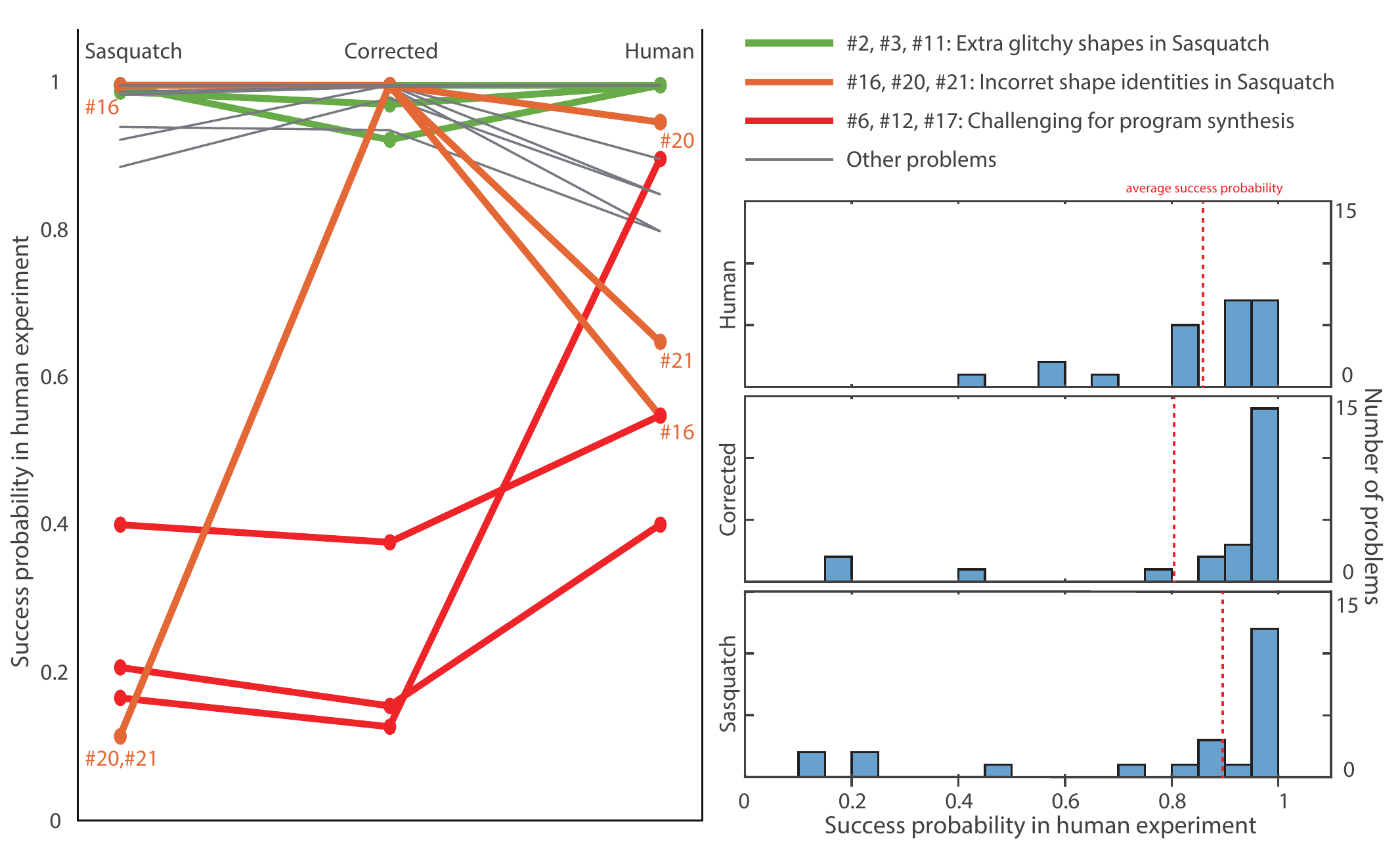}

\caption{\label{fig:Results-of-machine}The program synthesis performance with
the Sasquatch and our corrected parsings. The machine performance
$\alpha$ is reinterpreted with $\beta_{*}(\alpha)$ by Figure \ref{fig:markovchain}B.
The three histograms summarise the corresponding distributions of
$\beta$ and $\beta_{*}(\alpha)$, where the averages are $86.30\pm3.55\%$
for the human data, $81.40\pm7.01\%$ for the Sasquatch parsings and
$88.99\pm5.62\%$ for the corrected parsings. By modifying the representation
of image parsings, the machine performance were significantly enhanced
in problem \#20 and \#21. Although the machine performance achieved,
even surpassed, the human performance at many tasks, it was noticeably
poor in problems \#6, \#12 and \#17 (whose classification rule is
dependent on equidistance relations between shapes). The human performance
was calculated from the original data in \citet{fleuret2011comparing},
and the machine performance of Sasquatch was reproduced by the code
made public by \citet{ellis2015unsupervised} at \protect\url{https://github.com/ellisk42/sasquatch}. }
\end{figure}

By correcting the parsings, the performance of the PS classifier was
boosted from the chance level to nearly the perfect in problems \#20
and \#21, while it was similar in other problems (see Figure \ref{fig:Results-of-machine}
and Table \ref{tab:Summary-of-performance} for more details). The
PS performance is highly dependent on the parsings, which is one reason
for grouping the SVRT into different types as any parsing should in
principle encode all necessary information for classification (see
Section \ref{subsec:Grouping-SVRT}).

Other than the performance measured by the statistics ($\alpha$ or
$\beta_{*}(\alpha)$), we also consider the performance of the PS
classifier was improved in the sense of interpretability, which is
one of the main advantages of the PS approach. Similar to the discussion
on the AdaBoost performance, the PS classifier achieved decent performance
with both the Sasquatch and our parsings in problem \#16. However,
the synthesized programs in the two cases were different. With the
incorrect Sasquatch parsings, the program for the positive category
was to firstly draw a group of three shapes and then another group;
within either group the shapes were identical. With our corrected
parsings, the program was to draw six identical shapes with three
of them reflected.

Although the performance of the PS classifier was improved with the
corrected parsings, achieving even surpassing human performance in
many problems, it was relatively poor in problem \#17 and merely above
chance level in problem \#6 and \#12. The PS performance on these
three problems rely on the most complex core characteristics for classification
($LR=3$) as equidistance between shapes (a non-linear relation in
the shape coordinates) are involved. Although there was no constraint
for the PS classifier to access the coordinate information in the
parsings, its performance was inferior to the human level.

We did not attempt to re-engineer the parsings to encode the distance
information directly. Although the PS classifier could conduct linear
arithmetics over the distances rather than non-linear ones over the
coordinates, saving some computational expense, the re-engineering
would be at a huge cost in a combinatorial increase in the input sizes,
which the PS classifier could not afford because it scales poorly
with the input dimensionality. Moreover, even if the re-engineered
parsings had resulted in a decent performance in the three problems,
it could still be difficult for the classifier to generalise to unseen
problems, as they encode only the pieces of information according
to explicit human instructions. For example, we did not extract the
information of shape rotation for the parsings, because it is not
the core characteristics for classification in all the 23 problems.
Thus, it is straightforward to design a new SVRT problem \#101, which
would be unsolvable problem for the PS classifier due to the rotation
blindness of parsings, but not impossible for human subjects: all
images in both categories contain two shapes; in one category the
two shapes are identical, while in the other category the two shapes
become identical after rotating.

In summary, we corrected the parsings so that they contain in principle
all possible core characteristics for classification, which did lead
to some improvement in the PS performance on many, but not all, SVRT
problems.

\subsection{Challenging problems for humans and machines}

\begin{figure}
\centering

\includegraphics[width=1\textwidth]{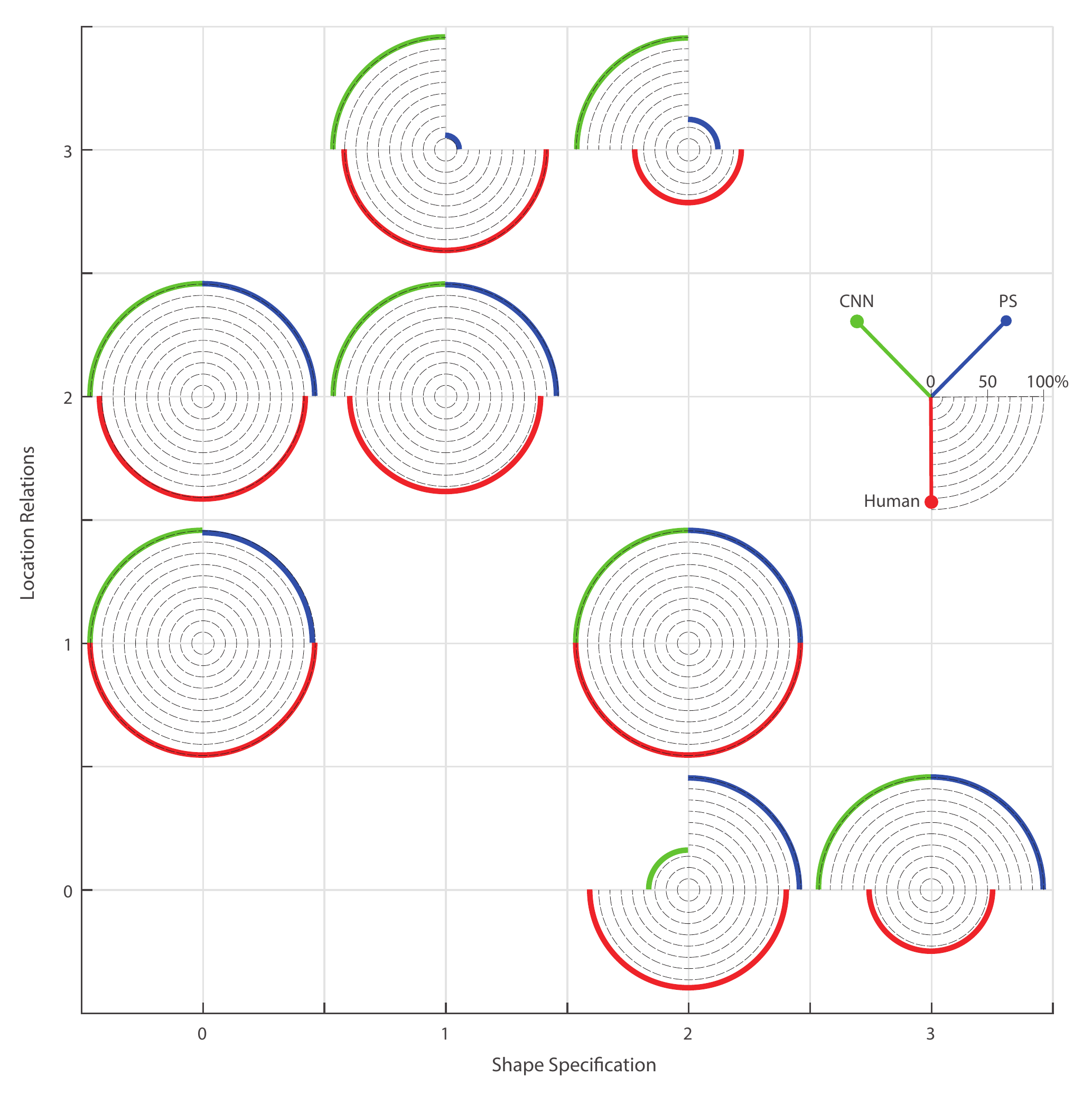}

\caption{\label{fig:Comparison-human-machine}Comparison between human and
machine (the best CNN and the PS classifier) performance on different
groups of the SVRT problems. The group performance is depicted in
propotion to the radii of the arcs.}
\end{figure}

Next, we compared the human and machine performance on the orignal
23 SVRT problems across the 8 different types to which they belong
(see Table \ref{tab:Different-types-23}). For compactness, we took
the best CNN performance reported in \citet{stabinger201625,ricci2018same}
as their performance were similar on most of the problems (see Table
\ref{tab:Summary-of-performance} for the detailed data). By averaging
the performance on different problems within each type, we obtained
human and machine performance ($\beta$ and $\beta_{*}(\alpha)$)
with respect to different levels of SS and LR as shown in Figure \ref{fig:Comparison-human-machine}.
In general, problem types on the left or lower in this $(SS,LR)$
plane are more complex than those on the right or upper in terms of
the core characteristics for classification. Although the humans,
the best CNN and the PS classifier with our corrected parsings actually
yielded great performance in most of the problem types, they exposed
their weaknesses when facing some problems types. The human performance
was relatively poor in $(SS,LR)=(2,3)$ and $(SS,LR)=(3,0)$. The
best CNN failed in $(SS,LR)=(2,0)$. The performance of the PS classifier
with the correct parsings in $(SS,LR)=(1,3)$ and $(SS,LR)=(2,3)$
was not significantly better than the chance level.

As discussed in Section \ref{subsec:Improved-PS-performance}, the
PS classifier failed to achieve a decent performance because it could
not detected the equidistance relations between shapes, whereas the
corrected parsings in principle provided it with the information to
calculate such relations. The problem types of $LR=3$ are challenging
to the PS classifier because non-linear arithmetics could be difficult
to its underlying theorem prover to verify.

The best CNN was worst in the problem type of $(SS,LR)=(2,0)$, which
means it is difficult for the CNNs to detect identical shapes within
an image. This limitation of CNNs was thoroughtly studied by \citet{ricci2018same}.
However, the performance was high in more difficult problem types
of $(SS,LR)=(3,0)$, $(2,1)$ and $(2,3)$. The great performance
in the type of $(SS,LR)=(3,0)$ was only due to LeNet, as its test
accuracy is nearly perfect while that is poor for GoogLeNet and vanilla
CNNs (see Table \ref{tab:Summary-of-performance}) \citep{stabinger201625,ricci2018same}.
In contrast, they all yielded decent performance in the problem types
of $(SS,LR)=(2,1)$ and $(2,3)$, specifically, on problem \#6, \#8
and \#17.

The human performance could be an evidence showing that it was sensible
for us to group the SVRT problems into these types. With an increase
in the levels of either SS or LR, it required a human to extract more
core characteristics for classification, making the problems more
difficult. Other than problem \#6, \#16 and \#17, which lie on the
most top and right on this $(SS,LR)$ plane and thus yielded the worst
3 human results out of the 23 problems, we also remark that problem
\#21 was the next most challenging problem for many humans. In contrast,
the human performance was nearly perfect on problem \#19 and \#20,
whereas the classification rules of these 3 problems are seemingly
alike in words and all belong to $(SS,LR)=(0,2)$. There are always
two shapes in an image, for the positive category the two shapes are
identical up to a similarity transformation, scaling for problem \#19,
reflection for problem \#20 and rotation for problem \#21, and for
the negative category the two shapes are simply different.

In summary, according to our grouping by SS and LR, we found that
finding non-linear relations in shape coordinates are the most challenging
to the PS classifier, detection of identical shapes to the CNNs, and
complex core characteristics for classification to the human subjects,
whereas some differences between problems within the same types might
have been overlooked.

\subsection{Few-shot learning by PS classifier}

\begin{figure}
\centering

\includegraphics[height=10cm]{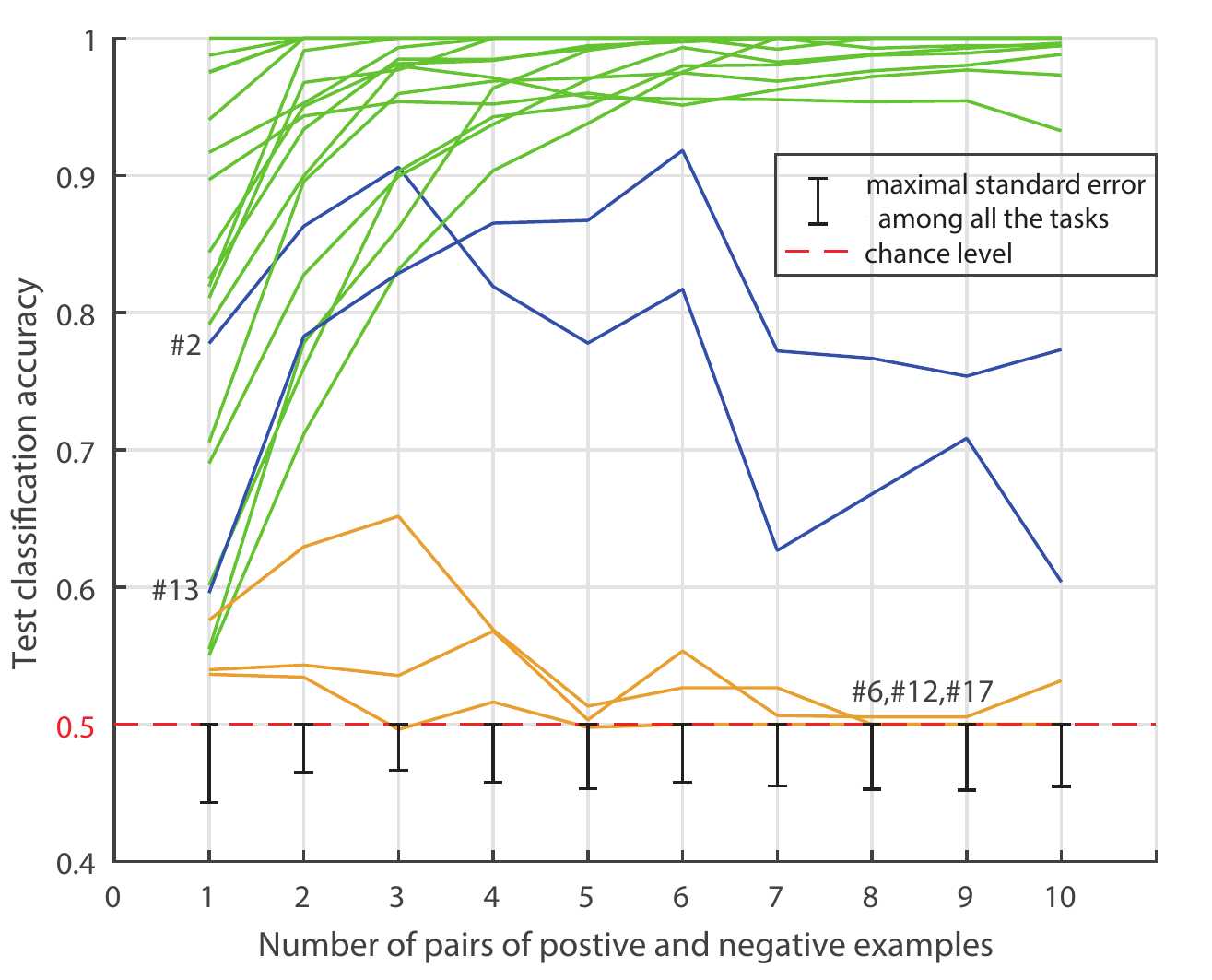}

\caption{\label{fig:learning-curve}Learning curves of the PS classifier for
the 23 tasks. The classifier reached optimal performance on many tasks
after learning a few examples (in green). However, noticeable declines
were observed for task \#2 and \#13 (in blue), which was caused by
incorrect image parsings. Moreover, the classifier struggled at about
chance level for task \#6, \#12, and \#17 regardless of the number
of the training examples (in orange). The classification rules of
these three tasks are all dependent on equidistance relations between
shapes, and the program synthesizer was unable to contruct optimal
programs encoding such relations. Standard errors were calculated
for all the data points, but only the maximal one for each size of
training sets is depicted to avoid messy plots. When the accuracies
approach $100\%$ (for many green curves), the corresponding standard
errors drop to $0\%$.}
\end{figure}

One advantage of the PS classifier is its capability of learning from
a small set of examples \citep{ellis2015unsupervised}; in particular,
data in the previous figures were obtained by testing the classifier
after training it with 3 positive and 3 negative examples. By varying
the number of pairs $t$, the test accuracy $\alpha(t)$ is effectively
the learning curve of the PS classifier, which is depicted in Figure
\ref{fig:learning-curve} for all the original 23 SVRT problems. Although
the PS classifier never achieved decent performance on problem \#6,
\#12 and \#17, it started to make correct classifications consistently
if not perfectly in many other problems after seeing no more than
6 pairs of positive and negative examples. Comparing to the other
ML techniques (e.g., AdaBoost, SVM and CNNs), the PS classifier needed
a extremely small training set.

However, a noticeable decline in the performance was observed for
problem \#2 and \#13. This behaviour of PS was essentially caused
by its formal verification nature . As any program can be synthesized
to instantiate all the training images, an insignificant but misleading,
random variability in the image parsings might be treated incorrectly
as the core characteristics for classification, which weakens the
program in generalising to unseen images. Moreover, if such misleading
variability is ever taken into account by PS, it becomes one of the
logical premises of the program. In other words, its impact cannot
be removed or reduced. With an increasing number of training examples,
the probability of at least one appearance of such random mistakes
grows rapidly, which makes the PS classifier less likely to achieve
high performance. In contrast, a statistical ML model (e.g., CNNs)
may suffer a deterioration in its performance due to overfitting,
which might also be a result of too powerful a model and too small
a training set. However, a statistical model is only penalised by
the deviation between its predication and the true label. As the frequency
of such misleading variability was presumbly constant, its impact
could be balanced out given more input data. In summary, more data
could be beneficial to the generalisation ability of a statistical
ML classifier. However, the same thing might be destructive to that
ability of a PS classifier, unless the data (the training image parsings)
are completely free of random mistakes.

\section{Discussion\label{sec:Discussion}}

In this paper we have proposed a quantitative reintepretation of the
human and machine performance on each SVRT problem for a fair comparison
despite the differences in the experimental protocols. We have also
grouped the problems into different types according to their core
characteristics for classification, so that we could analyse and compare
the human and machine performance on all the SVRT problems in a systematic
manner despite of the arbitrary design of the original problems.

We confirmed with AdaBoost that the classification based on the parsings
did not make the SVRT trivial, while incorrect parsings could make
a problem unsolvable. The PS performance was dependent on the quality
of the images parsings. With the corrected parsings, we improved the
general performance on the original SVRT problems. However, the PS
classifier still failed to detect equidistance relations in problem
\#6, \#12 and \#17, whereas it had access to the shape coordinates
and was able to construct a program including a command encoding distances
(computed from the coordinates). We also conducted the machine experiments
on these problems with the time limit of the synthesizer set to be
10 times and even 100 times larger than the original one. No improvement
in the performance was observed. We believe the failure is essentially
due to combinatorially complexity increase in the non-linear verification
in SMT. Movement is the only command in the synthesizer that is non-linear
in general because a displacement is calculated from a moving distance
and an orientation angle, which happens to be linear when there is
only one initial angle. Thus, the PS classifier could solve, for example,
problem \#13 but not \#12. It would also fail to generalise to other
problems involving distances according to our grouping of the problems,
unless a substaintially larger computational power could be deployed.

We grouped the problems based on the core characteristics for classification.
The complexity level of the core characteristics was qualitatively
ranked in the two aspects of SS and LR, while the complexity of the
real classification rules could be completely different. In fact,
the PS classifier chooses optimal programs by its evaluator measuring
the complexity of constructed programs. We did not group the problems
by the classification rules, because the core characteristics is more
objective. The complexity of programs is only defined for the PS classifier.
Humans might deduce different rules for the same category due to individual
differences, especially when the images appear complicated. It is
also not trivial for human to interpret a hyperplane found by AdaBoost
or SVM or latent variables of a well-trained CNN, whereas they are
virtually the classification rules to the machine agents. As the classification
rules had to be found from the core characteristics, it was expected
that the problems of higher SS and LR should on average yield no better
performance than those of a lower SS and LR. However, we noted that
the CNNs were better in the problem types of $(SS,LR)=(2,1)$ and
$(2,3)$ than $(2,0)$. One possible explanation could be the correlation
between SS and LR, whereas conceptually they might seem independent.
Imposing more or stronger constraints solely on the locations of identical
shapes might have made the entire image more regular, resulting in
a spatial pattern that could be more easily detected by the CNNs without
even noticing which shapes were identical. This argument would be
consistent with the point of view of \citet{ricci2018same}, as they
considered problem \#6 and \#17 to be purely about LR.

On the contrary, the recent success of training ResNet50 (a variant
of CNN) to a high performance on all the 23 SVRT problems by \citet{borowski2019notorious}
demonstrated the ability of feedforward convolutional architectures
to detect identical shapes, whereas some problem types might remain
difficult. Thus, in terms of classification accuracy, a well-trained
CNN is for now the best machine agent for the SVRT. However, decent
machine performance does not necessarily imply a human-like concept
being learnt by a machine \citep{borowski2019notorious}, while the
PS approach is intepretable to human as constructed programs are counterpart
to the classification rules, which might be associated to semantic
memory in human brain. Considering its capability of unsupervised,
few-shot learning, the PS approach is advantageous in some aspects,
whereas scalability is a major, general limitation. Due to this inherent
limitation, the PS performance has to be dependent on the quality
of image parsings.

One of the most fundamental differences between the PS and the statistical
ML classifier (e.g., AdaBoost, CNNs) is how much prior knowledge correlated
to the SVRT is manually encoded into the machine. We consider such
prior knowledge particularly worthy of discussion when comparing machine
and human performance and when investigating human-like computation
in machines. Despite the fact that the human subjects had never seen
the SVRT previously as the images were randomly generated, the human
subjects could hold some prior knowledge correlated to the SVRT before
the experiment, perhaps because the images involve higher-level configurations
that biological visiual systems can perceive effortlessly due to evolution
\citep{fleuret2011comparing}. In contrast, the statistical ML classifiers
were trained from scratch \citep{fleuret2011comparing,ellis2015unsupervised,stabinger201625,ricci2018same,borowski2019notorious},
holding little prior knowledge except that for the general purpose
of visual recognition. The SVRT images are known to consist of black
and white pixels only; AdaBoost and SVM were based on some feature
extractors; and all the CNNs deployed filters assuming translational
symmetry. Although transfer learning is a common recipe for filling
the gap of such prior knowledge to achieve few-shots learning in many
ML contexts \citep{pan2009survey,yosinski2014transferable}, it is
not a promising solution to the SVRT, because without careful choice
of model parameters they cannot perform decent classifications even
after being trained on thousands of examples in some problem. Much
more prior knowledge were manually encoded in the PS classifier. Although
it assumes that in general the synthesizer can construct a program
to instantiate any SVRT images in any category, it could be arguable
whether the classifier is specially engineered for the original 23
problems. Our grouping thus became relevant in the analysis, because
we could see which problem types, not individual problems, are trivial
but which are challenging to machines.

Although \citet{borowski2019notorious} showed that their machine
agent could solve all the 23 problems, the comparison between human
and machine performance remains a complicated issue, because the protocols
of the human and the machine experiments are different. Each human
subject was learning and tested at the same time througout the experiment,
while the machines are trained and tested separately. It is straightforward
to simulate how a machine agent, capable of online learning, makes
classifications as if it is in a human experiment. However, the issue
of prior knowledge would become more relevant as one would have to
decide to what degree the agent should be trained before implementing
transfer learning and online learning. For the PS classifier, its
poor performance on distance relations has to be fixed and a training
strategy preventing its fast performance decline needs to be deployed.

The PS classifier is advantageous for its capability of unsupervised,
few-shot learning, because the underlying theorem prover, Z3, performs
symbolic computation using high level representation of images (i.e.,
parsings). For the same reason, it has a scalability problem and relies
on the parsings, while the CNNs do not have such limitations. The
different advantages and disadvantages of the symbolic and neural
approaches suggests their combination might lead to greater machine
performance, e.g., \citet{ellis2018dreamcoder,huang2018gamepad,minervini2018towards,selsam2018learning}.

\bibliographystyle{plainnat}
\bibliography{references}

\newpage{}

\section{Appendix\label{appendix}}

\subsection{Summary of the SVRT problems: example images, classification rules
and problem types.\label{tab:Original-23}}

The original 23 SVRT problems were generated using our code, which
was forked from \citet{fleuret2011comparing}.

\begin{longtable}[c]{|>{\centering}m{1.5cm}|>{\centering}m{0.5cm}|>{\centering}p{4.5cm}|>{\raggedright}m{5cm}|c|}
\hline
Problem & \multicolumn{2}{l|}{Example} & Classification rule & $(SS,LR)$\tabularnewline
\hline
\endhead
\hline
\multirow{3}{1.5cm}{\#1} & \multicolumn{2}{>{\raggedright}m{4.5cm}|}{Basic feature} & There are two shapes. & \multirow{3}{*}{$(2,0)$}\tabularnewline
\cline{2-4}
 & + & %
{\fboxrule 0.2pt\fbox{\parbox[c][2cm]{2cm}{%
\includegraphics[width=1.8cm,height=1.8cm]{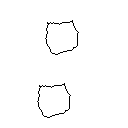}%
}}} & The two shapes are identical. & \tabularnewline
\cline{2-4}
 & - & %
{\fboxrule 0.2pt\fbox{\parbox[c][2cm]{2cm}{%
\includegraphics[width=1.8cm,height=1.8cm]{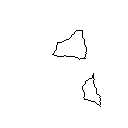}%
}}} & The two shapes are different. & \tabularnewline
\hline
\multirow{3}{1.5cm}{\#2} & \multicolumn{2}{l|}{Basic feature} & There are two shapes. The small shape is inside the large one. & \multirow{3}{*}{$(0,1)$}\tabularnewline
\cline{2-4}
 & + & %
{\fboxrule 0.2pt\fbox{\parbox[c][2cm]{2cm}{%
\includegraphics[width=1.8cm,height=1.8cm]{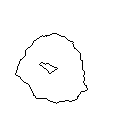}%
}}} & The small shape is near the centre of the large one. & \tabularnewline
\cline{2-4}
 & - & %
{\fboxrule 0.2pt\fbox{\parbox[c][2cm]{2cm}{%
\includegraphics[width=1.8cm,height=1.8cm]{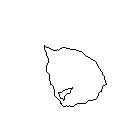}%
}}} & The small shape is near the boundary of the large one. & \tabularnewline
\hline
\multirow{3}{1.5cm}{\#3} & \multicolumn{2}{l|}{Basic feature} & There are four shapes. & \multirow{3}{*}{$(0,1)$}\tabularnewline
\cline{2-4}
 & + & %
{\fboxrule 0.2pt\fbox{\parbox[c][2cm]{2cm}{%
\includegraphics[width=1.8cm,height=1.8cm]{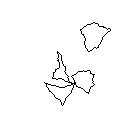}%
}}} & Three shapes are in contact. & \tabularnewline
\cline{2-4}
 & - & %
{\fboxrule 0.2pt\fbox{\parbox[c][2cm]{2cm}{%
\includegraphics[width=1.8cm,height=1.8cm]{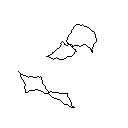}%
}}} & Within each pair, the two shapes are in contact. & \tabularnewline
\newpage
\hline
\multirow{3}{1.5cm}{\#4} & \multicolumn{2}{l|}{Basic feature} & There are two shapes. & \multirow{3}{*}{$(0,1)$}\tabularnewline
\cline{2-4}
 & + & %
{\fboxrule 0.2pt\fbox{\parbox[c][2cm]{2cm}{%
\includegraphics[width=1.8cm,height=1.8cm]{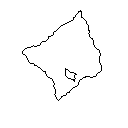}%
}}} & The small shape is inside the large one. & \tabularnewline
\cline{2-4}
 & - & %
{\fboxrule 0.2pt\fbox{\parbox[c][2cm]{2cm}{%
\includegraphics[width=1.8cm,height=1.8cm]{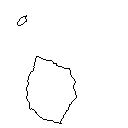}%
}}} & The small shape is outside the large one. & \tabularnewline
\hline
\multirow{3}{1.5cm}{\#5} & \multicolumn{2}{l|}{Basic feature} & There are four shapes. & \multirow{3}{*}{$(2,0)$}\tabularnewline
\cline{2-4}
 & + & %
\fbox{\parbox[c][2cm]{2cm}{%
\includegraphics[width=1.8cm,height=1.8cm]{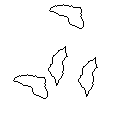}%
}} & There are two pairs of identical shapes. & \tabularnewline
\cline{2-4}
 & - & %
\fbox{\parbox[c][2cm]{2cm}{%
\includegraphics[width=1.8cm,height=1.8cm]{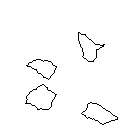}%
}} & The four shapes are different. & \tabularnewline
\hline
\multirow{3}{1.5cm}{\#6} & \multicolumn{2}{l|}{Basic feature} & There are two pairs of identical shapes. & \multirow{3}{*}{$(2,3)$}\tabularnewline
\cline{2-4}
 & + & %
{\fboxrule 0.2pt\fbox{\parbox[c][2cm]{2cm}{%
\includegraphics[width=1.8cm,height=1.8cm]{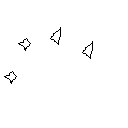}%
}}} & The distances within the pairs are the same. & \tabularnewline
\cline{2-4}
 & - & %
{\fboxrule 0.2pt\fbox{\parbox[c][2cm]{2cm}{%
\includegraphics[width=1.8cm,height=1.8cm]{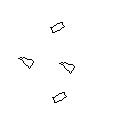}%
}}} & The distances within the pairs are random. & \tabularnewline
\hline
\multirow{3}{1.5cm}{\#7} & \multicolumn{2}{l|}{Basic feature} & There are six shapes. & \multirow{3}{*}{$(2,0)$}\tabularnewline
\cline{2-4}
 & + & %
{\fboxrule 0.2pt\fbox{\parbox[c][2cm]{2cm}{%
\includegraphics[width=1.8cm,height=1.8cm]{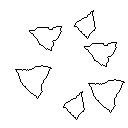}%
}}} & There are three groups of two identical shapes. & \tabularnewline
\cline{2-4}
 & - & %
{\fboxrule 0.2pt\fbox{\parbox[c][2cm]{2cm}{%
\includegraphics[width=1.8cm,height=1.8cm]{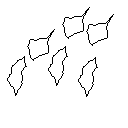}%
}}} & There are two groups of three identical shapes. & \tabularnewline
\newpage
\hline
\multirow{3}{1.5cm}{\#8} & \multicolumn{2}{l|}{Basic feature} & There are two shapes. & \multirow{3}{*}{$(2,1)$}\tabularnewline
\cline{2-4}
 & + & %
{\fboxrule 0.2pt\fbox{\parbox[c][2cm]{2cm}{%
\includegraphics[width=1.8cm,height=1.8cm]{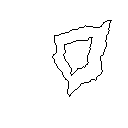}%
}}} & The small shape is inside the large one AND they are similar. & \tabularnewline
\cline{2-4}
 & - & %
{\fboxrule 0.2pt\fbox{\parbox[c][2cm]{2cm}{%
\includegraphics[width=1.8cm,height=1.8cm]{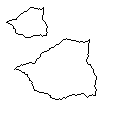}%
}}}%
{\fboxrule 0.2pt\fbox{\parbox[c][2cm]{2cm}{%
\includegraphics[width=1.8cm,height=1.8cm]{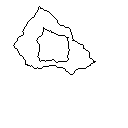}%
}}} & The small shape is outside the large one OR they are different. & \tabularnewline
\hline
\multirow{3}{1.5cm}{\#9} & \multicolumn{2}{l|}{Basic feature} & There are three shapes in a line. & \multirow{3}{*}{$(1,2)$}\tabularnewline
\cline{2-4}
 & + & %
{\fboxrule 0.2pt\fbox{\parbox[c][2cm]{2cm}{%
\includegraphics[width=1.8cm,height=1.8cm]{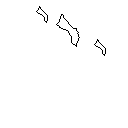}%
}}} & The large shape is in between the two small ones. & \tabularnewline
\cline{2-4}
 & - & %
{\fboxrule 0.2pt\fbox{\parbox[c][2cm]{2cm}{%
\includegraphics[width=1.8cm,height=1.8cm]{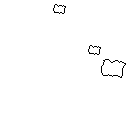}%
}}} & The large shape is on one end. & \tabularnewline
\hline
\multirow{3}{1.5cm}{\#10} & \multicolumn{2}{l|}{Basic feature} & There are four identical shapes. & \multirow{3}{*}{$(0,2)$}\tabularnewline
\cline{2-4}
 & + & %
{\fboxrule 0.2pt\fbox{\parbox[c][2cm]{2cm}{%
\includegraphics[width=1.8cm,height=1.8cm]{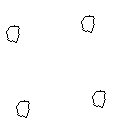}%
}}} & The shapes form a square. & \tabularnewline
\cline{2-4}
 & - & %
{\fboxrule 0.2pt\fbox{\parbox[c][2cm]{2cm}{%
\includegraphics[width=1.8cm,height=1.8cm]{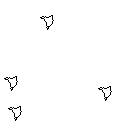}%
}}} & The shape locations are random. & \tabularnewline
\hline
\multirow{3}{1.5cm}{\#11} & \multicolumn{2}{l|}{Basic feature} & There are two shapes. & \multirow{3}{*}{$(0,1)$}\tabularnewline
\cline{2-4}
 & + & %
{\fboxrule 0.2pt\fbox{\parbox[c][2cm]{2cm}{%
\includegraphics[width=1.8cm,height=1.8cm]{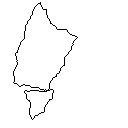}%
}}} & The two shapes are in contact. & \tabularnewline
\cline{2-4}
 & - & %
{\fboxrule 0.2pt\fbox{\parbox[c][2cm]{2cm}{%
\includegraphics[width=1.8cm,height=1.8cm]{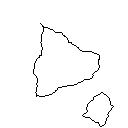}%
}}} & The two shapes are not in contact. & \tabularnewline
\newpage
\hline
\multirow{3}{1.5cm}{\#12} & \multicolumn{2}{l|}{Basic feature} & There are three shapes. & \multirow{3}{*}{$(1,3)$}\tabularnewline
\cline{2-4}
 & + & %
{\fboxrule 0.2pt\fbox{\parbox[c][2cm]{2cm}{%
\includegraphics[width=1.8cm,height=1.8cm]{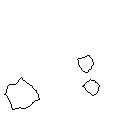}%
}}} & The two small shapes are close to each other.  & \tabularnewline
\cline{2-4}
 & - & %
{\fboxrule 0.2pt\fbox{\parbox[c][2cm]{2cm}{%
\includegraphics[width=1.8cm,height=1.8cm]{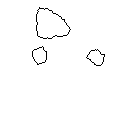}%
}}} & The shape locations are random. & \tabularnewline
\hline
\multirow{3}{1.5cm}{\#13} & \multicolumn{2}{l|}{Basic feature} & There are two identical large shapes and two identical small shapes. & \multirow{3}{*}{$(1,2)$}\tabularnewline
\cline{2-4}
 & + & %
{\fboxrule 0.2pt\fbox{\parbox[c][2cm]{2cm}{%
\includegraphics[width=1.8cm,height=1.8cm]{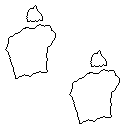}%
}}} & The two meta-shapes are identical; a meta-shape is a pair of a large
and a small shape. & \tabularnewline
\cline{2-4}
 & - & %
{\fboxrule 0.2pt\fbox{\parbox[c][2cm]{2cm}{%
\includegraphics[width=1.8cm,height=1.8cm]{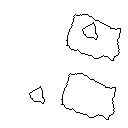}%
}}} & The shape locations are random. & \tabularnewline
\hline
\multirow{3}{1.5cm}{\#14} & \multicolumn{2}{l|}{Basic feature} & There are three identical shapes. & \multirow{3}{*}{$(0,2)$}\tabularnewline
\cline{2-4}
 & + & %
{\fboxrule 0.2pt\fbox{\parbox[c][2cm]{2cm}{%
\includegraphics[width=1.8cm,height=1.8cm]{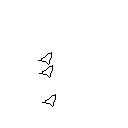}%
}}} & The shapes form a line. & \tabularnewline
\cline{2-4}
 & - & %
{\fboxrule 0.2pt\fbox{\parbox[c][2cm]{2cm}{%
\includegraphics[width=1.8cm,height=1.8cm]{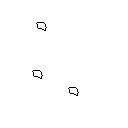}%
}}} & The shape locations are random. & \tabularnewline
\hline
\multirow{3}{1.5cm}{\#15} & \multicolumn{2}{l|}{Basic feature} & There are four shapes of the same size. They form a square. & \multirow{3}{*}{$(2,0)$}\tabularnewline
\cline{2-4}
 & + & %
{\fboxrule 0.2pt\fbox{\parbox[c][2cm]{2cm}{%
\includegraphics[width=1.8cm,height=1.8cm]{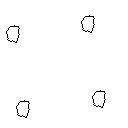}%
}}} & The four shapes are identical. & \tabularnewline
\cline{2-4}
 & - & %
{\fboxrule 0.2pt\fbox{\parbox[c][2cm]{2cm}{%
\includegraphics[width=1.8cm,height=1.8cm]{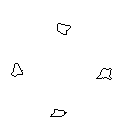}%
}}} & The four shapes are different. & \tabularnewline
\newpage
\hline
\multirow{3}{1.5cm}{\#16} & \multicolumn{2}{l|}{Basic feature} & There are six identical shapes. Their locations are symmetric with
respect to the vertical bisector of the image. & \multirow{3}{*}{$(3,0)$}\tabularnewline
\cline{2-4}
 & + & %
{\fboxrule 0.2pt\fbox{\parbox[c][2cm]{2cm}{%
\includegraphics[width=1.8cm,height=1.8cm]{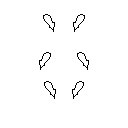}%
}}} & Three shapes are reflected. & \tabularnewline
\cline{2-4}
 & - & %
{\fboxrule 0.2pt\fbox{\parbox[c][2cm]{2cm}{%
\includegraphics[width=1.8cm,height=1.8cm]{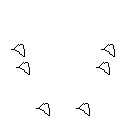}%
}}} & No shapes are reflected. & \tabularnewline
\hline
\multirow{3}{1.5cm}{\#17} & \multicolumn{2}{l|}{Basic feature} & There are four shapes, three of which are identical. & \multirow{3}{*}{$(2,3)$}\tabularnewline
\cline{2-4}
 & + & %
{\fboxrule 0.2pt\fbox{\parbox[c][2cm]{2cm}{%
\includegraphics[width=1.8cm,height=1.8cm]{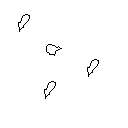}%
}}} & The distance between each of the three identical shapes and the different
one is the same. & \tabularnewline
\cline{2-4}
 & - & %
\fbox{\parbox[c][2cm]{2cm}{%
\includegraphics[width=1.8cm,height=1.8cm]{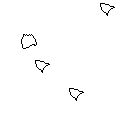}%
}} & The shape locations are random. & \tabularnewline
\hline
\multirow{3}{1.5cm}{\#18} & \multicolumn{2}{l|}{Basic feature} & There are six identical shapes. & \multirow{3}{*}{$(0,2)$}\tabularnewline
\cline{2-4}
 & + & %
{\fboxrule 0.2pt\fbox{\parbox[c][2cm]{2cm}{%
\includegraphics[width=1.8cm,height=1.8cm]{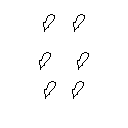}%
}}} & The shape locations are symmetric with respect to the vertical bisector
of the image. & \tabularnewline
\cline{2-4}
 & - & %
{\fboxrule 0.2pt\fbox{\parbox[c][2cm]{2cm}{%
\includegraphics[width=1.8cm,height=1.8cm]{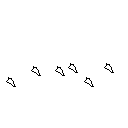}%
}}} & The shape locations are random. & \tabularnewline
\newpage
\hline
\multirow{3}{1.5cm}{\#19} & \multicolumn{2}{l|}{Basic feature} & There are two shapes of different sizes. & \multirow{3}{*}{$(2,0)$}\tabularnewline
\cline{2-4}
 & + & %
{\fboxrule 0.2pt\fbox{\parbox[c][2cm]{2cm}{%
\includegraphics[width=1.8cm,height=1.8cm]{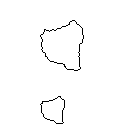}%
}}} & The two shapes are identical by scaling. & \tabularnewline
\cline{2-4}
 & - & %
{\fboxrule 0.2pt\fbox{\parbox[c][2cm]{2cm}{%
\includegraphics[width=1.8cm,height=1.8cm]{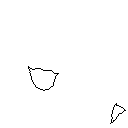}%
}}} & The two shapes are different. & \tabularnewline
\hline
\multirow{3}{1.5cm}{\#20} & \multicolumn{2}{l|}{Basic feature} & There are two shapes. & \multirow{3}{*}{$(2,0)$}\tabularnewline
\cline{2-4}
 & + & %
{\fboxrule 0.2pt\fbox{\parbox[c][2cm]{2cm}{%
\includegraphics[width=1.8cm,height=1.8cm]{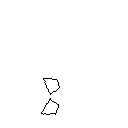}%
}}} & The two shapes are identical by reflection. & \tabularnewline
\cline{2-4}
 & - & %
{\fboxrule 0.2pt\fbox{\parbox[c][2cm]{2cm}{%
\includegraphics[width=1.8cm,height=1.8cm]{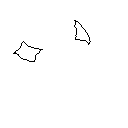}%
}}} & The two shapes are different. & \tabularnewline
\hline
\multirow{3}{1.5cm}{\#21} & \multicolumn{2}{l|}{Basic feature} & There are two shapes. & \multirow{3}{*}{$(2,0)$}\tabularnewline
\cline{2-4}
 & + & %
{\fboxrule 0.2pt\fbox{\parbox[c][2cm]{2cm}{%
\includegraphics[width=1.8cm,height=1.8cm]{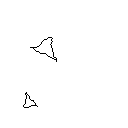}%
}}} & The two shapes are identical by scaling and rotation. & \tabularnewline
\cline{2-4}
 & - & %
{\fboxrule 0.2pt\fbox{\parbox[c][2cm]{2cm}{%
\includegraphics[width=1.8cm,height=1.8cm]{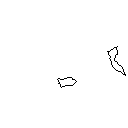}%
}}} & The two shapes are different. & \tabularnewline
\hline
\multirow{3}{1.5cm}{\#22} & \multicolumn{2}{l|}{Basic feature} & There are three shapes in a line. & \multirow{3}{*}{$(2,0)$}\tabularnewline
\cline{2-4}
 & + & %
{\fboxrule 0.2pt\fbox{\parbox[c][2cm]{2cm}{%
\includegraphics[width=1.8cm,height=1.8cm]{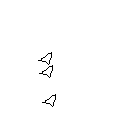}%
}}} & The shapes are identical. & \tabularnewline
\cline{2-4}
 & - & %
{\fboxrule 0.2pt\fbox{\parbox[c][2cm]{2cm}{%
\includegraphics[width=1.8cm,height=1.8cm]{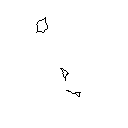}%
}}} & The three shapes are different. & \tabularnewline
\newpage
\hline
\multirow{3}{1.5cm}{\#23} & \multicolumn{2}{l|}{Basic feature} & There are two small and one large shapes. & \multirow{3}{*}{$(0,1)$}\tabularnewline
\cline{2-4}
 & + & %
{\fboxrule 0.2pt\fbox{\parbox[c][2cm]{2cm}{%
\includegraphics[width=1.8cm,height=1.8cm]{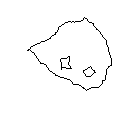}%
}}}%
{\fboxrule 0.2pt\fbox{\parbox[c][2cm]{2cm}{%
\includegraphics[width=1.8cm,height=1.8cm]{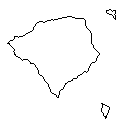}%
}}} & Both small shapes are inside OR outside the large one. & \tabularnewline
\cline{2-4}
 & - & %
{\fboxrule 0.2pt\fbox{\parbox[c][2cm]{2cm}{%
\includegraphics[width=1.8cm,height=1.8cm]{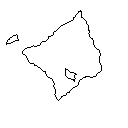}%
}}} & One small shape is inside the large one AND the other is outside. & \tabularnewline
\hline
\hline
\multirow{3}{1.5cm}{\#101} & \multicolumn{2}{l|}{Basic feature} & There are two shapes (up to rotation). & \multirow{3}{*}{$(3,0)$}\tabularnewline
\cline{2-4}
 & + & %
{\fboxrule 0.2pt\fbox{\parbox[c][2cm]{2cm}{%
\includegraphics[width=1.8cm,height=1.8cm]{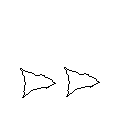}%
}}} & The two shapes are identical without rotation. & \tabularnewline
\cline{2-4}
 & - & %
{\fboxrule 0.2pt\fbox{\parbox[c][2cm]{2cm}{%
\includegraphics[width=1.8cm,height=1.8cm]{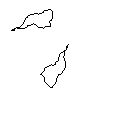}%
}}} & The two shapes are identical by rotation. & \tabularnewline
\hline
\end{longtable}

\subsection{Reinterpretation of machine performance with the success probability}

Since the statistical ML classifiers needs to be trained with thousands
of examples, they cannot perform classification and learning from
scratch simultaneously with only 35 images. We thus consider that
they have been well trained, and assume that, when participating this
human-like machine experiment, they do not update anymore. Under this
assumption, the experiment for a classifier of the test classification
accuracy $\alpha$ can be modelled as a Markov chain as shown in Figure
\ref{fig:markovchain}. It consists of $K+1$ states representing
$k\in\{0,1,2,\dots,\gamma-1,K\}$ correct classifications in a row.
The initial state is $k=0$, and $k=K$ is an absorbing state which
represents the success in the test. The transition probability from
the state $k$ to $k+1$ is $\alpha$ and that from $k$ to $0$ is
$1-\alpha$ for $k\neq K$. By defining $Q_{k}(N)$ to be the probability
of the state $k$ and $\mathbb{\mathbf{Q}}(N)=[Q_{0}(N),Q_{1}(N),Q_{2}(N),\cdots,Q_{K-1}(N),Q_{K}(N)]^{T}$
to be the probability distribution over $k\in\{0,1,2,\cdots,K-1,K\}$
at the $N$-th step, we can write down

\begin{equation}
\mathbf{Q}(N)=\mathbf{M}_{K}\mathbf{Q}(N-1),\label{eq:QNQN-1}
\end{equation}
where
\begin{equation}
\mathbf{M}_{K}=\left[\begin{array}{cccccc}
1-\alpha & 1-\alpha & \cdots & 1-\alpha & 1-\alpha & 0\\
\alpha & 0 & \cdots & 0 & 0 & 0\\
0 & \alpha & \cdots & 0 & 0 & 0\\
\vdots & \vdots & \ddots & \vdots & \vdots & \vdots\\
0 & 0 & \cdots & \alpha & 0 & 0\\
0 & 0 & \cdots & 0 & \alpha & 1
\end{array}\right]
\end{equation}
is the probability transition matrix. Therefore,

\begin{equation}
\mathbf{Q}(N)=\mathbf{M}_{K}^{N}\mathbf{Q}(0),
\end{equation}
where $\mathbb{\mathbf{Q}}(0)=[1,0,0,\cdots,0,0]^{T}$ is the initial
condition. In particular, $\beta_{*}(\alpha)=Q_{7}(35)$ is the probability
for a classifier to achieve a success using the human criterion of
\citet{fleuret2011comparing} (depicted in Figure \ref{fig:markovchain}).
Using this relationship, we can interpret the data of machine accuracy
$\alpha$ directly into machine success rate $\beta_{*}(\alpha)$
which is directly comparable to the human success rate $\beta$ (see
Table \ref{tab:Summary-of-performance}).

\begin{table}
\centering

\begin{tabular}{|c|c|c|c|c|c|c|c|}
\hline
\multirow{2}{*}{} & \multirow{2}{*}{Human} & \multicolumn{2}{c|}{PS} & \multicolumn{4}{c|}{CNN}\tabularnewline
\cline{3-8}
 &  & Sasquatch & Corrected & Best  & LeNet & GoogLeNet & Vanilla\tabularnewline
\hline
Task & $\beta$ & \multicolumn{3}{c|}{$\beta_{*}(\alpha)$} & \multicolumn{3}{c|}{$\alpha$}\tabularnewline
\hline
\hline
1 & 95\% & 100.00\% & 100.00\% & 33.88\% & 57\% & 50\% & 61.1\%\tabularnewline
\hline
2 & 100\% & 100.00\% & 92.51\% & 100.00\% & 100\% & 100\% & 100\%\tabularnewline
\hline
3 & 100\% & 99.17\% & 97.29\% & 100.00\% & N/A & N/A & 100\%\tabularnewline
\hline
4 & 100\% & 100.00\% & 100.00\% & 100.00\% & 100\% & 100\% & 100.0\%\tabularnewline
\hline
5 & 80\% & 98.59\% & 99.75\% & 46.24\% & 54\% & 50\% & 65.3\%\tabularnewline
\hline
6 & 40\% & 20.57\% & 15.51\% & 98.06\% & 76\% & 86\% & 87\%\tabularnewline
\hline
7 & 80\% & 94.17\% & 93.79\% & 22.82\% & 53\% & 50\% & 56.6\%\tabularnewline
\hline
8 & 100\% & 99.98\% & 99.97\% & 99.95\% & 94\% & 91\% & 93.4\%\tabularnewline
\hline
9 & 85\% & 100.00\% & 100.00\% & 100.00\% & 100\% & 100\% & 88.6\%\tabularnewline
\hline
10 & 95\% & 100.00\% & 100.00\% & 100.00\% & 99\% & 100\% & 100.0\%\tabularnewline
\hline
11 & 100\% & 100.00\% & 100.00\%  & 100.00\% & N/A & N/A & 100.0\%\tabularnewline
\hline
12 & 90\% & 16.56\% & 12.51\% & 100.00\% & 97\% & 100\% & 100.0\%\tabularnewline
\hline
13 & 85\% & 88.91\% & 98.09\%  & 99.32\% & N/A & N/A & 89.7\%\tabularnewline
\hline
14 & 95\% & 100.00\% & 99.99\% & 100.00\% & 90\% & 100\% & 96.1\%\tabularnewline
\hline
15 & 90\% & 100.00\% & 100.00\%  & 57.76\%  & 52\% & 50\% & 68.9\%\tabularnewline
\hline
16 & 55\% & 99.96\% & 100.00\% & 100.00\% & 98\% & 50\% & 76.5\%\tabularnewline
\hline
17 & 55\% & 40.06\% & 37.73\% & 99.98\% & 75\% & 95\% & 88.4\%\tabularnewline
\hline
18 & 85\% & 99.99\% & 99.94\% & 100.00\% & 99\% & 99\% & 100.0\%\tabularnewline
\hline
19 & 95\% & 92.59\% & 100.00\% & 30.94\% & 51\% & 50\% & 60.0\%\tabularnewline
\hline
20 & 95\% & 11.34\% & 100.00\% & 22.82\% & 55\% & 50\% & 56.6\%\tabularnewline
\hline
21 & 65\% & 11.34\% & 100.00\% & 28.15\% & 51\% & 51\% & 58.9\%\tabularnewline
\hline
22 & 100\% & 100.00\%  & 100.00\% & 37.24\% & 59\% & 50\% & 62.3\%\tabularnewline
\hline
23 & 100\% & 99.05\% & 99.67\% & 100.00\% & 87\% & 100\% & 93.2\%\tabularnewline
\hline
\end{tabular}

\caption{\label{tab:Summary-of-performance}Summary of the machine and human
performances on the original 23 SVRT problems. The human data came
from \citet{fleuret2011comparing}. The PS data was obtained by our
experiments. With the Sasquatch parsing, the experiment was a replication
of that in \citet{ellis2015unsupervised}. The LeNet and GoogLeNet
data came from \citet{stabinger201625}. The vanilla CNN data came
from \citet{ricci2018same}. The highest classification accuracy among
LeNet, GoogLeNet and vanilla CNN for each problem was chosen as the
best CNN accuracy $\alpha$, and was reinterpreted as success probability
$\beta_{*}(\alpha)$.}
\end{table}

\end{document}